\documentclass{article}

\usepackage[preprint]{neurips_2025}

\usepackage[utf8]{inputenc} 
\usepackage[T1]{fontenc}    
\usepackage{hyperref}       
\usepackage{url}            
\usepackage{booktabs}       
\usepackage{amsfonts}       
\usepackage{nicefrac}       
\usepackage{microtype}      
\usepackage{microtype}
\usepackage{graphicx}
\usepackage{booktabs} 
\usepackage{enumitem}
\usepackage{circledsteps}
\usepackage{colortbl}
\usepackage{hyperref}
\usepackage{algorithm}
\usepackage{algpseudocode}  

\usepackage{amsmath}
\usepackage{amssymb}
\usepackage{mathtools}
\usepackage{amsthm}
\usepackage{multirow}
\usepackage{diagbox}
\usepackage{graphicx}
\usepackage{subcaption}
\usepackage{booktabs}
\usepackage[capitalize,noabbrev]{cleveref}
\theoremstyle{plain}

\theoremstyle{definition}

\theoremstyle{remark}

\usepackage[textsize=tiny]{todonotes}

\title{
MedVista3D: Vision-Language Modeling for Reducing Diagnostic Errors in 3D CT Disease Detection, Understanding and Reporting
}

\author{%
    Yuheng Li \\
  Department of Biomedical Engineering\\
  Georgia Institute of Technology\\
  Atlanta, GA \\
  \And
  Yenho Chen \\
  Department of Machine Learning\\
  Georgia Institute of Technology\\
  Atlanta, GA \\
  \AND
  Yuxiang Lai \\
  Department of Radiation Oncology \\
  Emory University School of Medicine, Atlanta, USA \\
  Atlanta, GA \\
  \And
  Jike Zhong \\
  Department of Computer Science \\
  University of Southern California \\
  Los Angeles, CA \\
  \And
  Vanessa Wildman \\
  Department of Radiation Oncology \\
  Emory University School of Medicine, Atlanta, USA \\
  Atlanta, GA \\
  \And
  Xiaofeng Yang \\
  Department of Radiation Oncology \\
  Emory University School of Medicine, Atlanta, USA \\
  Atlanta, GA \\
  \texttt{xiaofeng.yang@emory.edu} \\
}

\begin{document}

\maketitle

\begin{abstract}
Radiologic diagnostic errors—under-reading errors, inattentional blindness, and communication failures—remain prevalent in clinical practice. These issues often stem from missed localized abnormalities, limited global context, and variability in report language. These challenges are amplified in 3D imaging, where clinicians must examine hundreds of slices per scan. Addressing them requires systems with precise localized detection, global volume-level reasoning, and semantically consistent natural language reporting. However, existing 3D vision-language models are unable to meet all three needs jointly—lacking local-global understanding for spatial reasoning and struggling with the variability and noise of uncurated radiology reports. We present MedVista3D, a multi-scale semantic-enriched vision-language pretraining framework for 3D CT analysis. To enable joint disease detection and holistic interpretation, MedVista3D performs local and global image-text alignment for fine-grained representation learning within full-volume context. To address report variability, we apply language model rewrites and introduce a Radiology Semantic Matching Bank for semantics-aware alignment. MedVista3D achieves state-of-the-art performance on zero-shot disease classification, report retrieval, and medical visual question answering, while transferring well to organ segmentation and prognosis prediction. Code and datasets will be released.

\end{abstract}

\section{Introduction}
Despite decades of clinical experience, radiologic diagnostic errors remain common and pose a persistent source of patient harm \cite{bruno2015understanding}. In a large-scale study \cite{kim2014fool}, three categories of errors were found to be prevalent. \textbf{Under-reading errors} occur when abnormalities are simply missed even within the field of view, often due to insufficient attention to localized findings. \textbf{Inattentional blindness} arise due to tunnel vision or limited global context, missing lesions outside the area of focus or in underexamined slices. \textbf{Communication failures} occur when correctly identified findings are ineffectively conveyed, often due to ambiguous phrasing or inconsistent terminology in the radiology report \cite{waite2018communication}. Addressing these errors requires systems capable of precise local detection, comprehensive image understanding, and clear, consistent communication of findings.

The development of such systems is particularly crucial for 3D medical images, where physicians must examine hundreds of cross-sectional slices which remains both time-consuming and expertise-driven \cite{shen2017deep}. Fundamentally, radiologic image interpretation spans three related tasks: (1) \textbf{localized detection} of anomalies like tumors or opacities; (2) \textbf{global understanding} of disease patterns across whole volume, which informs tasks like disease classification and report retrieval; and (3) \textbf{reporting}, which involves accurately describing findings and answering clinical questions in natural language. Recent advances in medical vision-language models (VLMs) have shown promise in automating these components—enabling localized disease identification, global image-report retrieval, zero-shot classification, and report generation or visual question answering \cite{fVLM, li2024artificial, thawkar2023xraygpt, zhang2023biomedgpt}.

\begin{figure*}[ht] 
    \centering
    \includegraphics[width=\textwidth]
    {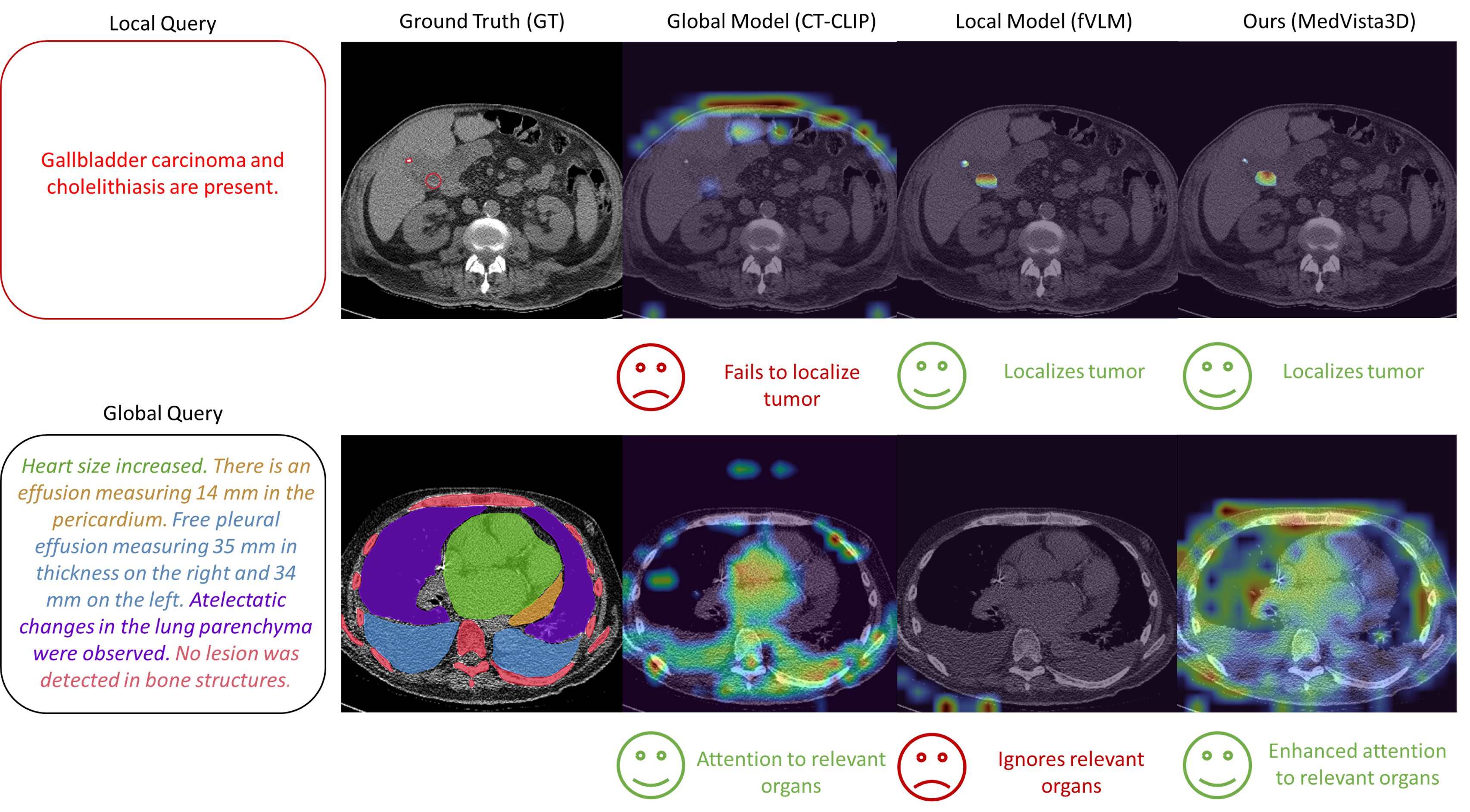}
    \vspace{-20pt}
    \caption{We visualize gradient activation maps for both global and local queries on global model (CT-CLIP) and local model (fVLM). Each row shows model attention for either a local (top) or global (bottom) query, with ground truth (GT) segmentations color-coded by sentence. The global model fails to detect tumor given a local query. The local model does not capture relevant anatomical regions given a global query. Our model effectively attends to relevant regions in both cases, demonstrating superior multi-scale understanding.}
    \label{fig:fig2}
\end{figure*}

\begin{figure}[ht] 
    \center
    \includegraphics[width=\textwidth]{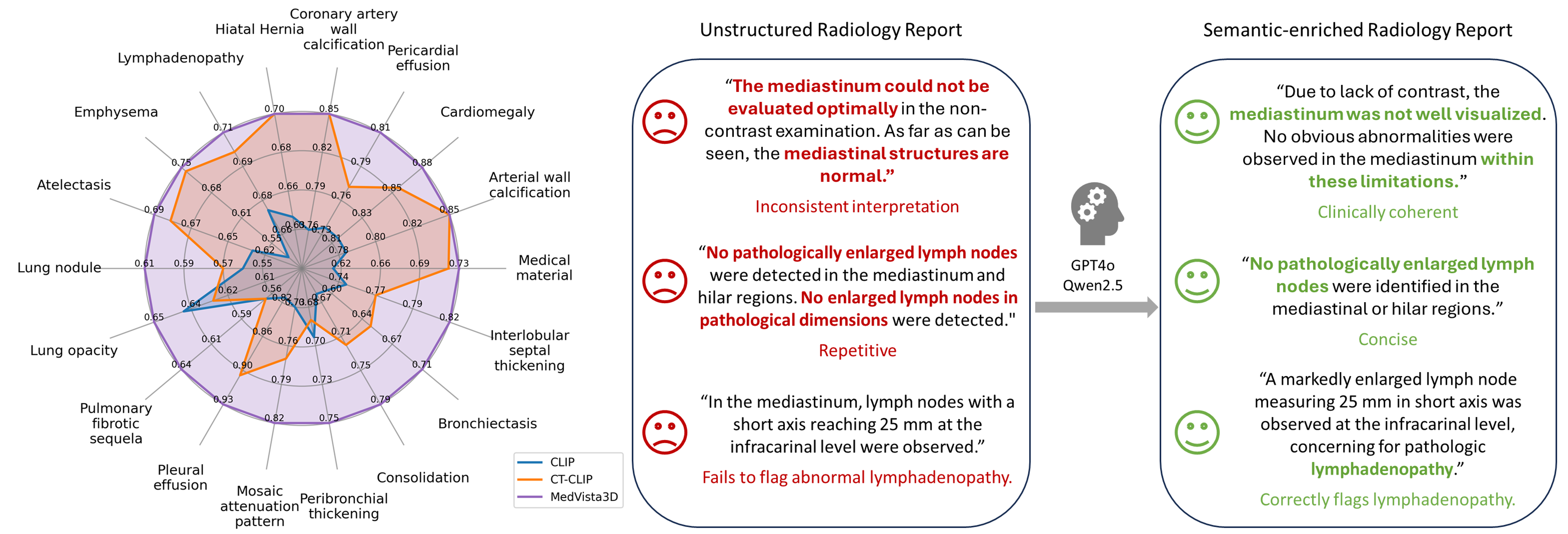} 
    \vspace{-10pt}
    \caption{
    \textbf{Left:} \textit{Global zero-shot performance of MedVista3D-ViT on CT-RATE.}
    AUC scores are reported per disease, reflecting the model's generalization across diverse pathologies.  
    \textbf{Right:} \textit{LLM-based refinement of radiology reports.}
    To address ambiguity and inconsistency in uncurated CT-RATE reports, we apply large language models (e.g., GPT-4o, Qwen2.5) to rewrite them with improved clarity and clinical coherence.
    }
    \label{fig:fig1}
    \vspace{-10pt}
\end{figure}

However, current medical VLMs cannot concurrently address these three diagnostic challenges, due to limitations in their training objectives and supervision data. \textbf{First}, existing models lack the capability to jointly perform local detection and global understanding, demonstrating under-reading and inattentional blindness for disease diagnosis. We analyze two state-of-the-art 3D CT VLMs (CT-CLIP \cite{hamamci2024developing} and fVLM \cite{fVLM}) under local and global settings for disease query. CT-CLIP is trained with a global-only objective, aligning entire volumes with full reports. As a result, it struggles to identify small and localized abnormalities due to insufficient local alignment. As shown in Figure~\ref{fig:fig2} (top row, second column), its gradient activations focus on irrelevant regions when queried about gallbladder carcinoma, paralleling the under-reading error. Conversely, fVLM aligns organ-level features with their corresponding text descriptions, but lacks a mechanism for global understanding. As shown in Figure~\ref{fig:fig2} (bottom row, third column), its activations neglect relevant organs under a global query, analogous to inattentional blindness where context beyond the region of focus is neglected. \textbf{Second}, the variability in real-world radiology reports could hinder learning consistent disease representations for effective reporting. Figure~\ref{fig:fig1} (right) shows text examples from a large-scale public dataset (e.g. CT-RATE). Empirically, we find that unstructured reports often contain inconsistent interpretations, repetitive phrasing, and vague expressions that fail to clearly convey clinically significant findings, such as lymphadenopathy. These issues degrade the quality of learned representations and introduce ambiguity in downstream tasks such as report generation and visual question answering (VQA). Addressing them requires medical VLMs to combine multi-scale visual grounding with semantically enriched alignment signals.


We propose \textbf{MedVista3D}, a 3D VLM to enhance the detection, understanding, and reporting of 3D CT image analysis. Our approach learns local-global representations while enhancing the disease-semantics understanding of the model. \textbf{First}, we derive a multi-scale loss that simultaneously aligns CT volumes and organ-level features with their corresponding text descriptions. This maximizes the mutual information shared between CT images and corresponding text descriptions, enabling both local detection and global understanding of the model. We theoretically demonstrate that this multi-scale loss captures more mutual information between global and local images and texts than single-scale alignment loss. A dual-pathway vision encoder is used to jointly process global 3D CT volumes and local segmented organs. \textbf{Second}, we improve semantic supervision through multi-scale semantic alignment. We enhance radiology reports via Large Language Model (LLM) rewrites to emphasize the presence or absence of each disease to ensure consistency. We then propose the Radiology Semantic Matching Bank (RSMB) for additional semantic alignment at global and local scales. RSMB retrieves semantically matched disease descriptions via nearest-neighbor search, providing robust text supervision. As shown in Figure~\ref{fig:fig1} (left), our MedVista3D considerably outperforms existing medical vision-language models on global disease zero-shot classification. 

We summarize the following contributions:
\begin{itemize}
    \item We identify the limitations of single-scale training objectives in existing 3D medical VLMs and derive a multi-scale alignment loss. We theoretically demonstrate that our loss can capture more shared cross-modal information than single-scale losses. Then, we present our unified objective and the corresponding architecture to jointly learn multi-scale representations from 3D CT volumes. 
    \item To address the variations in unstructured reports, we introduce multi-scale semantic supervision using LLM-rewritten reports and the Radiology Semantic Matching Bank, which retrieves semantically similar disease texts to enhance contrastive training across scales.
    \item We validate MedVista3D through comprehensive experiments across diverse medical tasks (e.g. \textit{disease zero-shot detection}, \textit{report retrieval}, \textit{medical VQA}, \textit{organ segmentation}, \textit{disease classification}), achieving state-of-the-art performance through unified alignment.
\end{itemize}
\section{Related work}
\textbf{Vision language models for medical imaging}. Previous medical VLMs predominantly employ global alignment, contrasting entire images and reports \cite{chauhan2020joint, zhang2024development, lin2023pmc, stevens2024bioclip, blankemeier2024merlin, hamamci2024developing}. More recent works introduced local region-text alignment \cite{ct-glip, fVLM} or local token-wise \cite{wang2022multi, huang2021gloria} alignment to learn fine-grained visual features. However, our investigation reveals a large gap between representations learned from global model and local model (Figure \ref{fig:fig2}). We motivate our approach by building a multi-scale pretraining method to combine the strengths of each alignment. 

\textbf{Multi-scale alignment for VLM.} There remains limited research in multi-scale alignment for VLMs. Existing methods \cite{huang2024enhancing, du2024multi} focus on multiscale radiography-report alignment but lack the use of region masks or bounding boxes for fine-grained detection. Other approaches \cite{chen2024contrastive, zhang2022glipv2} combine image-text and region-text alignments using coarse bounding boxes, which are less effective for precise organ localization. In contrast, our work utilizes segmentation masks to extract fine-grained organ features, enabling more accurate local alignment.

\textbf{Improving medical VLMs using synthetic data.} Given the scarcity of annotated data and privacy concerns in medical imaging, synthetic data has been widely explored to augment images \cite{koetzier2024generating,ozbey2023unsupervised,chlap2021review}. A few studies explored generating synthetic image or text data to support VLM pretraining \cite{wu2023medklip, liu2024can, bluethgen2024vision}. MedKlip \cite{wu2023medklip} extracts named entities from reports and supervises using these disease-specific queries. However, this approach overlooks the context and completeness of a query sentence. The closest to our work are local VLMs \cite{fVLM, ct-glip} that learn fine-grained representation using LLMs to decompose long reports into specific regions. However, these works do not learn multi-scale representations for local-global disease understanding, nor do they address the text variations in radiology reports. We tackle this by enhancing disease semantics in reports via LLMs and performing semantic alignment using nearest-neighbor search in the text embedding space.

\section{Method}
\begin{figure*}[ht] 
    \centering
    \includegraphics[width=\textwidth]{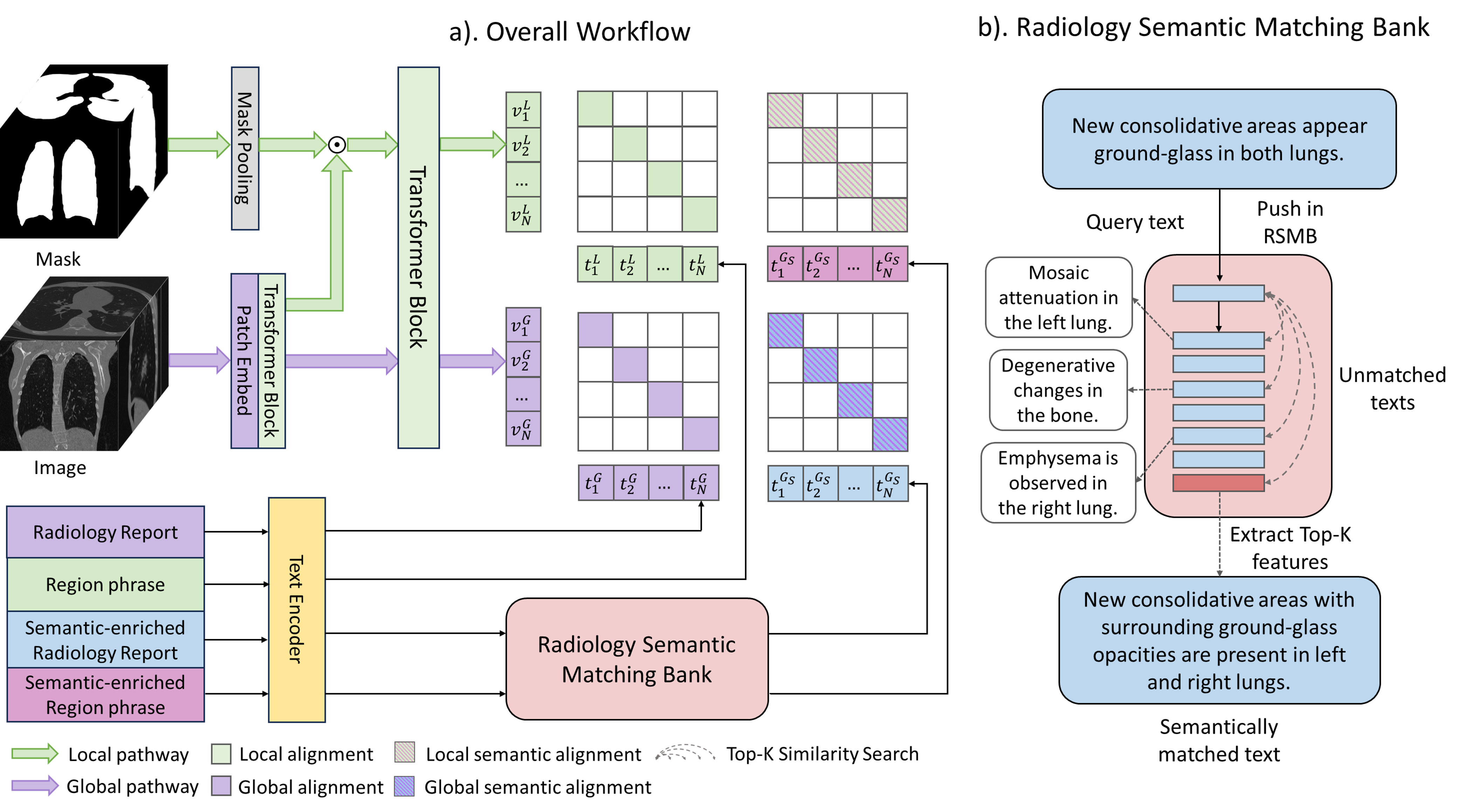} 
    \vspace{-10pt}
    \caption{a). MedVista3D encodes 3D CT volumes at both global and local scales. For local alignment, visual organ embeddings are paired with organ and semantic-enriched phrases. For global alignment, global volume embedding is matched with the report embedding and its semantic-enriched versions augmented by LLMs. b). A radiology semantic matching bank maintains a queue of text embeddings from diverse radiology descriptions. For each query, a top-k similarity search retrieves semantically matching texts, filtering out less relevant ones.}
    \vspace{-10pt}
    \label{fig:workflow}
\end{figure*}

\textbf{Overview.} Our approach performs alignment at four levels (Figure~\ref{fig:workflow}): (1) global volume with report, (2) local region with text, (3) global volume with a semantically enriched report, and (4) local region with a semantically enriched sentence. The multi-scale alignment strategy is detailed in Section~3.1, while semantic alignment using LLM-based rewrites and the Radiology Semantic Matching Bank is presented in Section~3.2. The global and local vision-language models used in our framework are described in Appendix~C.

\subsection{A Multi-scale Global and Local Alignment}
\textbf{Connection to mutual information maximization (MI).} Contrastive vision-language pretraining can be viewed as maximizing the mutual information between the image and text. MI quantifies the shared information between two random variables by measuring how much knowing one variable reduces the uncertainty about the other. Given CT volume as \(X_G\) and paired report as \(Y_G\), global alignment maximizes mutual information between full-volume and full-report pairs, $I_{G}(X_G; Y_G)$, where $(x_g, y_g)$ are volume-report pairs. Similarly, given CT region as \(X_L\) and region text as \(Y_L\), local alignment maximizes mutual between organ regions and their descriptions, $I(X_L; Y_L )$, where $(x_r, y_r)$ are region-text pairs and $R$ is the total number of regions defined. From \cite{poole2019variational}, contrastive loss (InfoNCE) estimates a lower bound for MI. Extending this theorem to both global and local alignment, we have: 
\begin{equation}
I(X_G; Y_G) \geq  - \mathcal{L}_{\rm \text{Global}} + \log(N_G), \label{eq:global_bound}
\end{equation}
\begin{equation}
I(X_L; Y_L ) \geq - \mathcal{L}_{\rm \text{Local}} + \log(N_L), \label{eq:local_bound}
\end{equation}
where $N_G$ and $N_L$ are the number of negative samples in global and local alignment, respectively. $\mathcal{L}_{\rm \text{Global}}$ and $\mathcal{L}_{\rm \text{Local}}$ are defined in equation \ref{eq:dual_global} and equation \ref{eq:3} respectively. 

\textbf{A multi-scale local and global objective.} We observe that equations~\refeq{eq:global_bound} and~\refeq{eq:local_bound} can only capture partial structure as they focus exclusively on either local or global views separately. Our core insight is that capturing both the holistic contexts and fine-grained details requires maximizing a unified mutual information between the full set of global and local CT images $X=(X_L, X_G)$ and text reports $Y=(Y_L, Y_G)$, defined as,
\begin{equation}
\begin{aligned}
I_{\text{Unified}}(X, Y) &= I(X_G, X_L; Y_L, Y_G). \\
\end{aligned}
\end{equation}
By the chain rule for mutual information, we have $I_{\text{Unified}}(X, Y) \geq \max{\{I(X_L;Y_L), I(X_G;Y_G)}\}$, indicating that the unified objective can capture more shared information between the modalities than considering either global or local inputs alone. This makes it better suited for learning representations that encode both global semantics and local alignment.
However, directly optimizing $I_{\text{Unified}}(X, Y)$ is computationally intractable. Instead, we propose a multi-scale contrastive loss that linearly integrates global and local alignment:
\begin{equation}
\mathcal{L_{\text{Multi-scale}}} = \frac{1}{2} \left[\mathcal{L_{\text{Global}}} + \mathcal{L}_{\text{Local}} \right]. \label{eq:Multi-scale}
\end{equation}
This objective is part of a valid lower bound,
\begin{equation}
\begin{aligned}
I_{\text{Unified}}(X, Y) &\geq -\mathcal{L_{\text{Multi-scale}}} + \frac{1}{2}\left[ \log (N_L) + \log (N_G)\right],\\
\end{aligned}
\end{equation}
and explicitly encourages the learned representation to jointly capture information from both global and local views of the input data.






\textbf{Architecture details.} Our MedVista3D is designed with 2 pathways (Figure \ref{fig:workflow}a). For global pathway, given CT volume $x_i$, the model first extracts patch embeddings $p_i \in \mathbb{R}^{c \times d \times h \times w}$ using a 3D convolutional layer. Transformer blocks then generate a latent image embedding $v_i^{{L}}$, which the final transformer block refines into the global image embedding $v_i^{G}$. The radiology report is encoded by text encoder into a global text embedding $t_i^{G}$. For local pathway, given anatomical region $r$ and its segmentation map $M_i^r \in \{0, 1\}^{D \times H \times W}$, mask pooling downsamples it to $\tilde{M}_i^r \in \{0, 1\}^{d \times h \times w}$, matching the patch grid resolution. Active region tokens (threshold 0.5 in $\tilde{M}_i^r$) are selected from $p_i$ element-wise. These are processed by the last transformer block to produce the local image embedding $v_i^r$. The corresponding region-specific text phrase is encoded into a local text embedding $t_i^{r}$. This preserves the spatial relationships between global and local embeddings.

\subsection{Radiology Semantic Enrichment and Alignment}

\textbf{Semantic enrichment of radiology reports.} While multi-scale alignment yields richer representations, communication errors could still be caused by directly training on unstructured reports. Free-form radiology reports often suffer from length and inconsistent terminologies (e.g., \textit{nodular opacities} vs. \textit{lesions}). To address this, we prompt the LLMs to identify all possible abnormalities from the report and rewrite the findings as discrete, presence-or-absence statements. This process, applied to both global reports and local region phrases, yields standardized, succinct text descriptions where each sentence details at least one abnormality (prompts in Appendix D).

\textbf{Radiology semantics matching bank.} Building on these enriched texts, the RSMB provides robust supervision by retrieving semantically similar embeddings, addressing minor wording variations. We observe that enriched texts often describe the same findings with only minor variations in wording (e.g., “mild pleural thickening” vs. “slight pleural thickening”). RSMB is a 64k-sized first-in-first-out queue storing previously encoded enriched global and local text features. For a new enriched query text, its top-1 nearest neighbor (via cosine similarity) is retrieved from RSMB. The corresponding image embedding is then aligned with this retrieved text, ensuring robustness to text variations while maintaining consistent disease semantics.

\textbf{Multi-scale semantic alignment.} Using RSMB and the enriched texts, we establish semantic alignment at two levels. For global-level, a semantically-enriched text embedding $\hat{t}_i^{G}$ queries RSMB to retrieve its nearest neighbor $\hat{t}_i^{G_{NN}}$. This embedding is aligned with global image embedding $v_i^{G}$ using contrastive loss: 
\begin{align}
\mathcal{L}_{\text{Global Semantic}} &= \mathcal{L}^{G_{NN}}_{I \to T} + \mathcal{L}^{G_{NN}}_{T \to I}, \label{eq:5}
\end{align}
\begin{align}
\mathcal{L}^{G_{NN}}_{I \to T} = -\frac{1}{N} \sum_{i=1}^N 
\log \frac{\exp \Big( \text{sim}(v_i^{G}, \hat{t}_i^{G_{NN}}) / \tau \Big)}
{\sum_{j=1}^N \exp \Big( \text{sim}(v_i^{G}, \hat{t}_j^{G_{NN}}) / \tau \Big)}.
\end{align}
Similarly for local-level, a semantically-enriched region embedding $\hat{t}_{i}^{r}$ queries RSMB for its neighbor $\hat{t}_i^{r_{NN}}$, which is aligned with the region image embedding $v_i^{r}$: 
\begin{align} 
\mathcal{L}_{\text{Local Semantic}} &= \mathcal{L}^{L_{NN}}_{I \to T} + \mathcal{L}^{L_{NN}}_{T \to I}, \label{eq:7} 
\end{align}
\begin{align}
\mathcal{L}^{L_{NN}}_{I \to T} = -\frac{1}{RN} \sum_{r=1}^R \sum_{i=1}^N 
\log \frac{\exp \Big( \text{sim}\big(v_i^{r}, \hat{t}_i^{r_{NN}})\big) / \tau \Big)}
{\sum_{j=1}^N \exp \Big( \text{sim}\big(v_i^{r}, \hat{t}_j^{r_{NN}}\big) / \tau \Big)}
\end{align}
We combine global and local alignment through,
\begin{equation}
\mathcal{L}_{\text{Multi-scale Semantic}} = \mathcal{L}_{\text{Global Semantic}}+ \mathcal{L}_{\text{Local Semantic}}.
\end{equation}
Combining both multi-scale and semantic alignment, our final pretraining objective is defined as,
\begin{equation}
\begin{split}
\mathcal{L}_{\text{MedVista3D}} = \mathcal{L}_{\text{Multi-scale}}+\mathcal{L}_{\text{Multi-scale Semantic}}. 
\end{split}
\label{eq:9}
\end{equation}

\section{Experiments} 
\textbf{Pretraining dataset and implementation.}
We pretrain MedVista3D on the CT-RATE dataset \cite{hamamci2024developing} using the training split (24,128 volumes) and perform testing on the internal test split (1,564 volumes). Local alignment uses Radgenome masks and region texts \cite{radgenome}. The vision encoder is either a ViT \cite{metaclip} or UniMISS \cite{unimiss}, and the text encoder is a pretrained BERT \cite{boecking2022making}. See Appendix B for details.

\begin{table*}[h!]
\caption{Performance comparison of VLMs at global tasks and local task on CT-RATE. \textcolor{blue!70}{\textbf{Blue}} for global models and \textcolor{green!70}{\textbf{green}} for local model. \textbf{BOLD} means best result and \underline{underline} second best. \dag\ : our implementation. \ddag\ : using official checkpoint. }
\centering
\small
\resizebox{\textwidth}{!}{%
\begin{tabular}{lcccccccccc}
\toprule
\multirow{3}{*}{Method} 
& \multicolumn{6}{c}{Global} 
& \multicolumn{4}{c}{Local} \\
\cmidrule(lr){2-7} \cmidrule(lr){8-11}
& \multicolumn{4}{c}{Disease zero-shot} 
& \multicolumn{2}{c}{Report retrieval} 
& \multicolumn{4}{c}{Disease zero-shot} \\
\cmidrule(lr){2-5} \cmidrule(lr){6-7} \cmidrule(lr){8-11}
& Precision & F1 & ACC & AUC 
& Recall 5 & Recall 10 
& Precision & F1 & ACC & AUC \\
\midrule
\rowcolor{blue!10} CLIP \cite{radford2021learning} &
0.334	& 0.726	& 0.691	& 0.703	& 2.67\%	& 5.00\%
& 0.306 & 0.696 & 0.657 & 0.659 \\
\rowcolor{blue!10} Merlin\textsuperscript{\ddag} \cite{blankemeier2024merlin}
& 0.229 & 0.612 & 0.558 & 0.578 
& 1.11\% & 2.02\% 
& 0.199 & 0.479 & 0.433 & 0.538 \\
\rowcolor{blue!10} CT-CLIP \cite{hamamci2024developing} 
& 0.306 & 0.691 & 0.651 & 0.704 
& 2.34\% & 3.95\% 
& 0.297 & 0.678 & 0.636 & 0.645 \\
\midrule
\rowcolor{green!10} fVLM\textsuperscript{\dag} \cite{fVLM}
& 0.293 & 0.684 & 0.641 & 0.644 
& 1.82\% & 3.06\% 
& 0.372	& 0.752	& 0.722	& 0.759\\
\rowcolor{green!10} fVLM\textsuperscript{\ddag} \ \cite{fVLM}
& 0.248 & 0.684 & 0.600 & 0.591
& 0.32\% & 1.09\%
& \textbf{0.379} & 0.751 & 0.718 & \underline{0.778} \\
\midrule
MedVista3D-ViT (ours)
& \underline{0.379}	& \underline{0.760}	& \underline{0.737}	& \underline{0.778}
& \textbf{6.64\%}	& \textbf{10.68\%}
& \underline{0.377}	& \textbf{0.765}	& \textbf{0.742}	& \textbf{0.780} \\
MedVista3D-UniMISS (ours)
& \textbf{0.385}	& \textbf{0.770}	& \textbf{0.745}	& \textbf{0.782}	
& \underline{5.01\%}	& \underline{8.65\%}	
& 0.372	& \underline{0.754}	& \underline{0.726}	& 0.753 \\
\bottomrule
\end{tabular}
}
\vspace{-10pt}
\label{tab:compare_CTRATE_table}
\end{table*}

\begin{table*}[h!]
\caption{Generalization of VLMs at global tasks and local task on Rad-ChestCT. \textcolor{blue!70}{\textbf{Blue}} for global models and \textcolor{green!70}{\textbf{green}} for local model. \textbf{BOLD} means best result and \underline{underline} second best. \dag\ : our implementation. \ddag\ : using official checkpoint.}
\centering
\small
\resizebox{\textwidth}{!}{%
\begin{tabular}{lcccccccc}
\toprule
\multirow{3}{*}{Method} 
& \multicolumn{4}{c}{Global} 
& \multicolumn{4}{c}{Local} \\
\cmidrule(lr){2-5} \cmidrule(lr){6-9}
& \multicolumn{4}{c}{Disease zero-shot} 
& \multicolumn{4}{c}{Disease zero-shot} \\
\cmidrule(lr){2-5} \cmidrule(lr){6-9}
& Precision & F1 & ACC & AUC 
& Precision & F1 & ACC & AUC \\
\midrule
\rowcolor{blue!10} CLIP \cite{radford2021learning}
& 0.352	& 0.637	& 0.617	& 0.609
& 0.321 &	0.593 & 0.569 & 0.559 \\
\rowcolor{blue!10} Merlin\textsuperscript{\ddag} \cite{blankemeier2024merlin}
& 0.339	& 0.605	& 0.581	& 0.596
& 0.210	& 0.562	&0.513	&0.552\\
\rowcolor{blue!10} CT-CLIP \cite{hamamci2024developing} 
&0.339	&0.648	&0.599	&0.632 
&0.334 & 0.608 & 0.584 & 0.689 \\
\midrule
\rowcolor{green!10} fVLM\textsuperscript{\dag} \cite{fVLM}
& 0.314	& 0.587	& 0.562	& 0.518
& 0.315	&0.596	&0.571	&0.524\\
\rowcolor{green!10} fVLM\textsuperscript{\ddag} \ \cite{fVLM}
&0.332	&0.561	&0.535	&0.544 
&0.374	&\textbf{0.688}	&\underline{0.647}	&0.680 \\
\midrule
MedVista3D-ViT (ours)
&\textbf{0.426}	&\textbf{0.693}	&\textbf{0.684}	&\underline{0.702}
&\textbf{0.402}	&\underline{0.681}	&\textbf{0.668}	&\textbf{0.710}\\
MedVista3D-UniMISS (ours)
&\underline{0.393}	&\underline{0.664}	&\underline{0.646}	& \textbf{0.713}
& \underline{0.378}	& 0.650	& 0.628	& \underline{0.697} \\
\bottomrule
\end{tabular}
}
\vspace{-10pt}
\label{tab:compare_radchestct}
\end{table*}

\begin{table}[ht]
\centering
\caption{Comparison of various LLaVA architectures on medical VQA (long answer, short answer, report generation, and multiple choice) on CT-RATE. \textbf{BOLD} means best result and \underline{underline} second best. \textcolor{blue!70}{\textbf{Blue}} for 2D MLLMs and \textcolor{green!70}{\textbf{green}} for 3D MLLMs.}
\resizebox{\textwidth}{!}{%
\begin{tabular}{lcccccccc}
\toprule
\multirow{2}{*}{\cellcolor{white}Method} 
& \multicolumn{4}{c}{Long Answer} 
& \multicolumn{4}{c}{Short Answer} \\
\cmidrule(lr){2-5} \cmidrule(lr){6-9}
& BLEU\_1 & METEOR & ROUGE\_L & CIDER 
& BLEU\_1 & METEOR & ROUGE\_L & CIDER \\
\midrule
\rowcolor{blue!10} CXR-LLaVA    
& 0.203 & 0.140 & 0.231 & 0.577 
& 0.016 & 0.000 & 0.021 & 0.040 \\
\rowcolor{blue!10} LLaVA-Med    
& 0.137 & 0.156 & 0.202 & 0.315 
& 0.014 & 0.051 & 0.025 & 0.007 \\
\rowcolor{green!10} CT-CHAT     
& \underline{0.480} & \underline{0.294} & \underline{0.512} & \underline{3.100} 
& \underline{0.280} & \underline{0.160} & \underline{0.598} & {\bf 1.821} \\
\rowcolor{green!10} MedVista3D-LLaVA (ours)      
& {\bf 0.516} & {\bf 0.309} & {\bf 0.546} & {\bf 3.395} 
& {\bf 0.299} & {\bf 0.178} & {\bf 0.602} & \underline{1.817} \\
\bottomrule
\end{tabular}
}

\vspace{1em} 

\resizebox{\textwidth}{!}{%
\begin{tabular}{lcccccccc}
\toprule
\multirow{2}{*}{\cellcolor{white}Method} 
& \multicolumn{4}{c}{Report Generation} 
& \multicolumn{4}{c}{Multiple Choice} \\
\cmidrule(lr){2-5} \cmidrule(lr){6-9}
& BLEU\_1 & METEOR & ROUGE\_L & CIDER 
& BLEU\_1 & METEOR & ROUGE\_L & CIDER\\
\midrule
\rowcolor{blue!10} CXR-LLaVA    
& 0.050 & 0.000 & 0.020 & 0.049
& 0.057 & 0.009 & 0.063 & 0.065\\
\rowcolor{blue!10} LLaVA-Med    
& 0.002 & 0.024 & 0.056 & 0.000
& 0.085 & 0.175 & 0.135 & 0.151\\
\rowcolor{green!10} CT-CHAT     
& \underline{0.381} & \underline{0.217} & \underline{0.334} & \underline{0.221} 
& \underline{0.838} & \underline{0.578} & \underline{0.895} & \underline{7.850}\\
\rowcolor{green!10} MedVista3D-LLaVA (ours)       
& {\bf 0.474} & {\bf 0.252} & {\bf 0.386} & {\bf 0.349} 
& {\bf 0.936} & {\bf 0.668} & {\bf 0.927} & {\bf 8.210}\\
\bottomrule
\end{tabular}
}
\vspace{-10pt}
\label{tab:llava}
\end{table}

\subsection{Reducing under-reading and inattentional blindness via MedVista3D}
We evaluate how MedVista3D reduces under-reading and inattentional blindness by assessing both global understanding and local disease detection from CT volumes. We compare with global VLMs—trained on the entire CT volume and corresponding text (e.g., CLIP \cite{radford2021learning}, CT-CLIP \cite{hamamci2024developing}, Merlin \cite{blankemeier2024merlin})—and local VLMs aligning region features with region text (e.g., fVLM \cite{fVLM}). 

\textbf{Local Task}: Assess under-reading errors by evaluating localized disease detection within anatomical regions (lungs, heart, aorta, and esophagus).
\begin{enumerate}[nosep,leftmargin=*]
  \item \emph{Disease zero-shot classification:} Given text prompts and segmentation masks, identify the presence of diseases. For global models we crop the CT volume to the segmentation mask and perform padding. Metrics include area under the ROC curve (AUC), balanced accuracy (ACC), precision, and weighted F1-score (F1).
\end{enumerate}

\textbf{Global Tasks}: Address inattentional blindness by evaluating the model’s ability to detect findings outside expected regions and retrieve reports correctly describing relevant findings.
\begin{enumerate}[nosep,leftmargin=*]
  \item \emph{Disease zero-shot classification:} Given text prompts, identify the presence of diseases in the CT volume without segmentation masks. We report the same metrics as in the local task. 
  \item \emph{Report retrieval:} Given a CT volume, retrieve the corresponding radiology report from the entire dataset. We measure recall at top-5 and top-10.
\end{enumerate}

\textbf{Results for local task.} On localized zero-shot detection, both MedVista3D backbones match or surpass fVLM. Importantly, fVLM suffers from poor generalization to global tasks (AUC drops from 0.759 to 0.644), whereas MedVista3D maintains superior performance across both tasks. This demonstrates our model’s ability to reduce under-reading errors. 

\textbf{Results for global tasks.} Both MedVista3D-ViT and MedVista3D-UniMISS outperform all global models in disease zero-shot and report retrieval (Table \ref{tab:compare_CTRATE_table}). For global disease zero-shot, MedVista3D-UniMISS achieves the highest AUC (0.782) and F1 (0.770), outperforming CT-CLIP by 7.4 points in AUC and 6.9 points in F1. For report retrieval, MedVista3D-ViT surpasses CT-CLIP by 4.3\% and 6.7\% in top-5 and top-10 recall. These results validate our model’s ability to jointly reduce inattentional blindness and under-reading errors.

\textbf{External validation.}
To assess generalization, we perform external validation on the full Rad-ChestCT dataset \cite{radchestct} (3626 volumes), following CT-CLIP and fVLM. We evaluate on global and local zero-shot disease detection. Segmentation masks are obtained using TotalSegmentator model \cite{wasserthal2023totalsegmentator}. As shown in Table \ref{tab:compare_radchestct}, MedVista3D-ViT consistently outperforms existing global models (CLIP, Merlin, CT-CLIP) across all global metrics, achieving an AUC of 0.702. MedVista3D-UniMISS achieves the highest global AUC of 0.713. For local tasks, MedVista3D-ViT also surpasses the fVLM with an AUC of 0.710, demonstrating robust generalization capabilities.

\begin{figure*}[ht] 
    \centering
    \includegraphics[width=\textwidth]
    {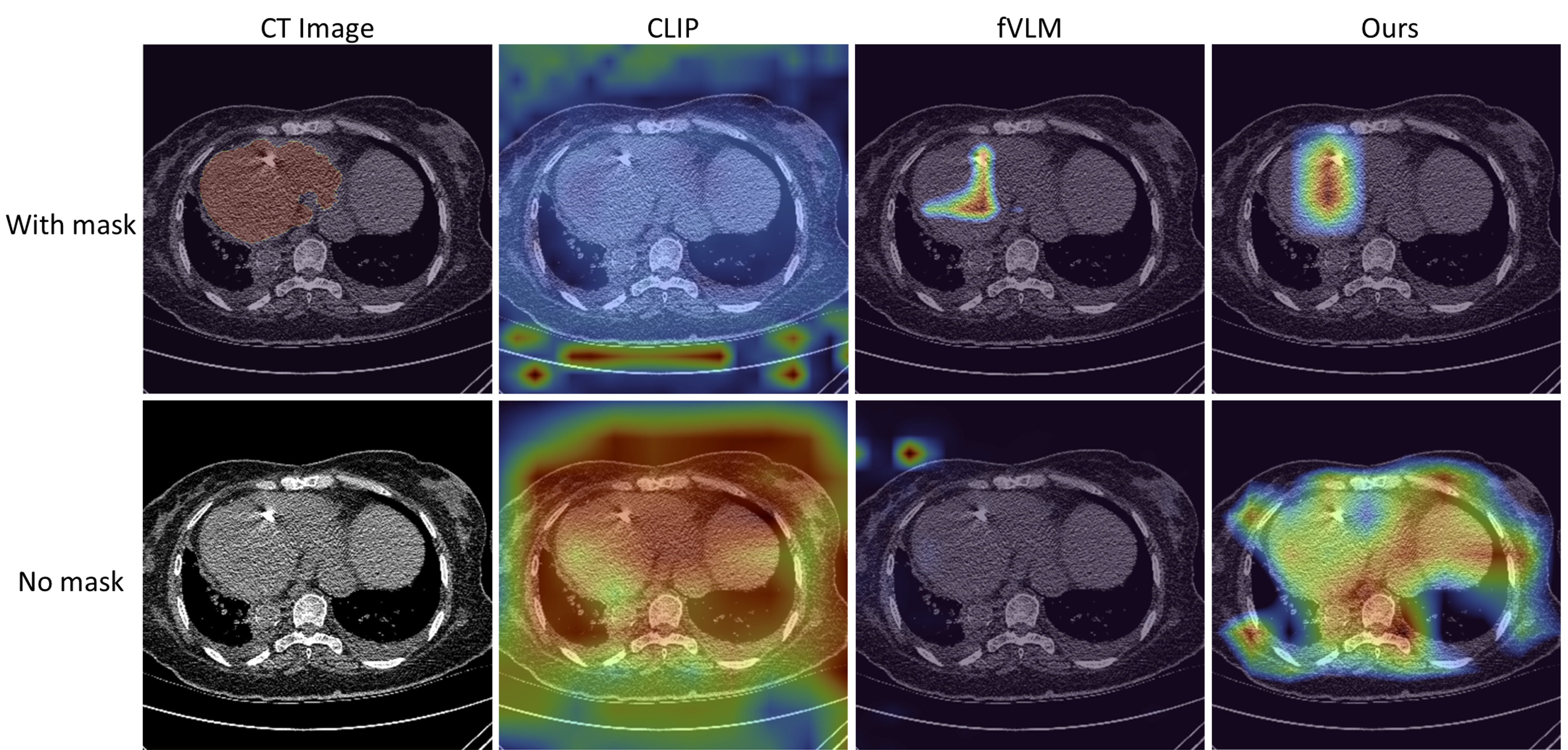}
    \vspace{-10pt}
    \caption{Impact of region masking on attention for CLIP, fVLM and MedVista3D (on CT-RATE). We visualize the attention maps of [CLS] token with other patch tokens given CT volume with (top) and without (bottom) region mask. MedVista3D remains focused on important organs regardless of masking. With mask CLIP shows diffuse attention; fVLM struggles without the mask.}
    \vspace{-10pt}
    \label{fig:attention}
\end{figure*}

\textbf{Qualitative results.} Figure \ref{fig:attention} illustrates how region masking modulates the attention map of [CLS]-to-patch tokens in CLIP, fVLM, and our model. CLIP shows broad attention without masking, but it fails to localize the correct region with mask, reflecting under-reading and explaining its poor local detection performance. fVLM attends correctly with a mask but fixates on irrelevant, tiny background areas without it, indicating inattentional blindness and poor global understanding.  In contrast, MedVista3D demonstrates both fine-grained attention with mask and global attention on the anatomy without mask, effectively mitigating both error types.

\subsection{Mitigating communication errors with MedVista3D-LLaVA}
To evaluate how our model mitigates communication errors in CT reporting, we train MedVista3D-LLaVA, a multimodal large language model (MLLM), on the CT-RATE VQA dataset. The dataset includes long-answer questions, short-answer questions, multiple-choice questions, and report generation tasks. Following our pretraining setup, we train on the CT-RATE training split and validate on its internal validation split. Evaluation follows CT-CHAT \cite{hamamci2024developing}, using BLEU, METEOR, ROUGE\_L, and CIDER scores. As shown in Table \ref{tab:llava}, our method consistently outperforms CT-CHAT as well as 2D multimodal assistants (LLaVA-Med \cite{Llava-med}, CXR-LLaVA \cite{CXR-LLAVA}) by considerable margins. It achieves the best performance on multiple-choice questions (BLEU\_1: 0.936, METEOR: 0.668, ROUGE\_L: 0.927, CIDER: 8.21), and surpasses CT-CHAT by 3.6\% and 1.9\% BLEU\_1 on long and short answer tasks, respectively. On the accuracy of multiple choice, our method achieves 91.5\%. For report generation, our model shows a 9.3-point BLEU\_1 improvement.  These gains demonstrate the effectiveness of our multi-scale alignment and semantic enrichment of radiology reports, which mitigate potential communication errors in diagnostic workflows.

\begin{table*}[h]
\centering
\caption{Ablation study on multi-scale and semantic image-text alignment.}
\resizebox{1\textwidth}{!}{%
\begin{tabular}{l c c c c c c c c}
\toprule
\multirow{2}{*}{Pretraining Strategy} &
\multicolumn{2}{c}{Region phrase grounding} &
\multicolumn{2}{c}{Report retrieval} &
\multicolumn{4}{c}{Global disease zero-shot} \\
\cmidrule(lr){2-3} \cmidrule(lr){4-5} \cmidrule(lr){6-9}
 & Top 10 & Top 50 & Top 5 & Top 10 & Precision & F1 & ACC & AUC \\
\midrule
Global Alignment      & 0.04\%  & 0.36\%    & 4.53\% & 7.88\% & 0.293 & 0.689 & 0.633 & 0.675 \\
+ Local Alignment  & 0.19\%  & 0.76\%  & 4.98\% & 8.32\% & 0.281 & 0.676 & 0.634 & 0.664 \\
+ Mask Pooling       & 0.48\%  & 2.42\%  & 4.53\% & 7.88\% & 0.279 & 0.674 & 0.631 & 0.609 \\
+ Global Semantic Alignment & 0.38\% & 1.99\% & 4.98\% & 8.25\% & 0.398 & 0.789 & 0.758 & 0.807 \\
+ Local Semantic Alignment  & 0.83\%  & 3.46\%  & 6.64\% & 10.68\% & 0.379 & 0.760 & 0.737 & 0.778 \\
\bottomrule
\end{tabular}%
}
\vspace{-10pt}
\label{tab:ablation_study}
\end{table*}

\subsection{Ablation Study}
\textbf{Ablation on multi-scale and semantic alignment.} We conduct a detailed analysis on our proposed objective loss on both global and local tasks using CT-RATE. For local task, we evaluate region phrase grounding, where the model retrieves the correct region description given an image and a segmentation mask. As summarized in Table \ref{tab:ablation_study}, we begin with a single global-level loss, with moderate global disease zero-shot and report retrieval performance, but lacking local capabilities. We further add local alignment which enables region grounding by explicitly learning organ-level embeddings. Next, we add mask pooling to allow more focused vision features on the segmentation mask, which further improve the region grounding but slightly compromising global zero-shot performance. Adding our proposed semantic alignment at global level considerably boosts global disease zero-shot and maintaining decent region retrieval. Finally, adding local semantic alignment yields the best overall balance between global and local tasks. 

\textbf{Ablation on mask pooling.} We conduct a study on the mask pooling mechanism of our method on local disease zero-shot. Specifically, we choose various layers of vision transformer blocks to perform organ mask pooling. As shown in Table \hyperref[tab:masking_depth]{5}, we find that applying mask pooling before the last transformer block yields the best performance. Applying mask pooling either before the first block or before the second block yields suboptimal performance. 

\begin{table}[htbp]
  \centering
  \begin{subtable}[t]{0.49\linewidth}
    \centering
    \caption*{Table 5: Ablation study on transformer block for mask pooling.}
    \begin{tabular}{@{}lcccc@{}}
      \toprule
      \multicolumn{1}{c}{} & \multicolumn{4}{c}{Local disease zero-shot} \\
      \cmidrule(lr){2-5}
      Layer & Precision & F1 & ACC & AUC \\
      \midrule
      1\textsuperscript{st} block & 0.336  & 0.709  & 0.656  & 0.703  \\
      2\textsuperscript{nd} block & 0.342  & 0.716  & 0.666  & 0.732  \\
      12\textsuperscript{th} block & 0.377  & 0.765  & 0.742  & 0.780  \\
      \bottomrule
    \end{tabular}
  \end{subtable}
  \hfill
  \begin{subtable}[t]{0.49\linewidth}
    \centering
    \caption*{Table 6: Organ segmentation on TotalSegmentator and prognosis prediction on STOIC.}
    \begin{tabular}{@{}lcc@{}}
      \toprule
      Method & TotalSeg (DSC) & STOIC (AUC) \\
      \midrule
      nnUNet & 0.852 & - \\
      CT-CLIP & 0.805 & 0.631 \\
      Merlin & \underline{0.860} & \underline{0.782} \\
      M3D & 0.597 & 0.627\\
      RadFM & - & 0.649 \\
      Ours & \textbf{0.872} & \textbf{0.807} \\
      \bottomrule
    \end{tabular}
  \end{subtable}
  \vspace{-10pt}
\label{tab:masking_depth}
\label{tab:downstream_cls_seg}
\end{table}
\subsection{Additional Applications}
\textbf{Organ Segmentation.} MedVista3D learns transferable representations for organ segmentation. We finetune our model using Totalsegmentator \cite{wasserthal2023totalsegmentator} which contains 1204 patients and 104 organs, covering a wide range of anatomical structures. We attach a U-Net decoder to our MedVista3D-UniMISS backbone for adaptation to segmentation task. We evaluate the segmentation performance using dice coefficient (DSC). We use nnUNet's default 5-fold cross-validation split for training and testing. MedVista3D-UniMISS achieves a DSC of 0.872 on TotalSegmentator, outperforming the state-of-the-art nnUNet by 2 points in DSC and Merlin by 1.2 point (Table \hyperref[tab:downstream_cls_seg]{6}).

\textbf{Prognosis prediction.} MedVista3D also enables accurate COVID prognosis prediction. We finetune using STOIC 2021 \cite{stoic} dataset for pneumonia severity prediction. We randomly select 80\% for training, 10\% for validation and 10\% for testing. A linear head is attached for classifying severe or non-severe (defined as death or need for intubation). AUC is used to evaluate the performance. MedVista3D-UniMISS achieves 0.807 AUC outperforming all comparable methods. These results show the adaptability of our method beyond multi-modal tasks.


\section{Conclusion}

We present \textbf{MedVista3D}, a 3D VLM using multi-scale semantic alignment to address three major diagnostic errors in radiology: under-reading, inattentional blindness, and communication failures. To jointly support local detection and global understanding from 3D CT volumes, we propose a multi-scale alignment loss based on mutual information maximization. To mitigate variability in report language, we leverage LLM-based rewrites and introduce a Radiology Semantic Matching Bank for robust semantic alignment. MedVista3D consistently outperforms existing 3D VLMs across multiple downstream tasks, including zero-shot disease classification, report retrieval, and VQA. It also demonstrates strong transferability to organ segmentation and prognosis prediction, highlighting its potential as a general-purpose foundation model for 3D medical imaging.

\textbf{Limitation and Future Work.} Our current pretraining is limited to chest CT, primarily due to the lack of large-scale, publicly available 3D image-report datasets for other anatomical regions. In future work, we aim to expand MedVista3D to include additional anatomical sites such as the brain, head-and-neck region, and pelvis. Moreover, we plan to extend our framework to other imaging modalities, such as MRI and PET, to further enhance its generalizability across clinical contexts.

\bibliographystyle{plain}
\bibliography{reference}
\newpage
\section*{NeurIPS Paper Checklist}

\begin{enumerate}

\item {\bf Claims}
    \item[] Question: Do the main claims made in the abstract and introduction accurately reflect the paper's contributions and scope?
    \item[] Answer: \answerYes{}
    \item[] Justification: The claims made in the abstract and introduction are aligned with the actual contributions and findings presented in the paper. Each of these claims is supported by theoretical analysis, ablation studies, and experimental results.
    \item[] Guidelines:
    \begin{itemize}
        \item The answer NA means that the abstract and introduction do not include the claims made in the paper.
        \item The abstract and/or introduction should clearly state the claims made, including the contributions made in the paper and important assumptions and limitations. A No or NA answer to this question will not be perceived well by the reviewers. 
        \item The claims made should match theoretical and experimental results, and reflect how much the results can be expected to generalize to other settings. 
        \item It is fine to include aspirational goals as motivation as long as it is clear that these goals are not attained by the paper. 
    \end{itemize}

\item {\bf Limitations}
    \item[] Question: Does the paper discuss the limitations of the work performed by the authors?
    \item[] Answer: \answerYes{} 
    \item[] Justification: See conclusion section.
    \item[] Guidelines:
    \begin{itemize}
        \item The answer NA means that the paper has no limitation while the answer No means that the paper has limitations, but those are not discussed in the paper. 
        \item The authors are encouraged to create a separate "Limitations" section in their paper.
        \item The paper should point out any strong assumptions and how robust the results are to violations of these assumptions (e.g., independence assumptions, noiseless settings, model well-specification, asymptotic approximations only holding locally). The authors should reflect on how these assumptions might be violated in practice and what the implications would be.
        \item The authors should reflect on the scope of the claims made, e.g., if the approach was only tested on a few datasets or with a few runs. In general, empirical results often depend on implicit assumptions, which should be articulated.
        \item The authors should reflect on the factors that influence the performance of the approach. For example, a facial recognition algorithm may perform poorly when image resolution is low or images are taken in low lighting. Or a speech-to-text system might not be used reliably to provide closed captions for online lectures because it fails to handle technical jargon.
        \item The authors should discuss the computational efficiency of the proposed algorithms and how they scale with dataset size.
        \item If applicable, the authors should discuss possible limitations of their approach to address problems of privacy and fairness.
        \item While the authors might fear that complete honesty about limitations might be used by reviewers as grounds for rejection, a worse outcome might be that reviewers discover limitations that aren't acknowledged in the paper. The authors should use their best judgment and recognize that individual actions in favor of transparency play an important role in developing norms that preserve the integrity of the community. Reviewers will be specifically instructed to not penalize honesty concerning limitations.
    \end{itemize}

\item {\bf Theory assumptions and proofs}
    \item[] Question: For each theoretical result, does the paper provide the full set of assumptions and a complete (and correct) proof?
    \item[] Answer: \answerYes{}
    \item[] Justification: See section 3.1 and 3.2.
    \item[] Guidelines:
    \begin{itemize}
        \item The answer NA means that the paper does not include theoretical results. 
        \item All the theorems, formulas, and proofs in the paper should be numbered and cross-referenced.
        \item All assumptions should be clearly stated or referenced in the statement of any theorems.
        \item The proofs can either appear in the main paper or the supplemental material, but if they appear in the supplemental material, the authors are encouraged to provide a short proof sketch to provide intuition. 
        \item Inversely, any informal proof provided in the core of the paper should be complemented by formal proofs provided in appendix or supplemental material.
        \item Theorems and Lemmas that the proof relies upon should be properly referenced. 
    \end{itemize}

    \item {\bf Experimental result reproducibility}
    \item[] Question: Does the paper fully disclose all the information needed to reproduce the main experimental results of the paper to the extent that it affects the main claims and/or conclusions of the paper (regardless of whether the code and data are provided or not)?
    \item[] Answer: \answerYes{}
    \item[] Justification: See Appendix B for detailed descriptions of model architecture, training settings, hyperparameter.
    \item[] Guidelines:
    \begin{itemize}
        \item The answer NA means that the paper does not include experiments.
        \item If the paper includes experiments, a No answer to this question will not be perceived well by the reviewers: Making the paper reproducible is important, regardless of whether the code and data are provided or not.
        \item If the contribution is a dataset and/or model, the authors should describe the steps taken to make their results reproducible or verifiable. 
        \item Depending on the contribution, reproducibility can be accomplished in various ways. For example, if the contribution is a novel architecture, describing the architecture fully might suffice, or if the contribution is a specific model and empirical evaluation, it may be necessary to either make it possible for others to replicate the model with the same dataset, or provide access to the model. In general. releasing code and data is often one good way to accomplish this, but reproducibility can also be provided via detailed instructions for how to replicate the results, access to a hosted model (e.g., in the case of a large language model), releasing of a model checkpoint, or other means that are appropriate to the research performed.
        \item While NeurIPS does not require releasing code, the conference does require all submissions to provide some reasonable avenue for reproducibility, which may depend on the nature of the contribution. For example
        \begin{enumerate}
            \item If the contribution is primarily a new algorithm, the paper should make it clear how to reproduce that algorithm.
            \item If the contribution is primarily a new model architecture, the paper should describe the architecture clearly and fully.
            \item If the contribution is a new model (e.g., a large language model), then there should either be a way to access this model for reproducing the results or a way to reproduce the model (e.g., with an open-source dataset or instructions for how to construct the dataset).
            \item We recognize that reproducibility may be tricky in some cases, in which case authors are welcome to describe the particular way they provide for reproducibility. In the case of closed-source models, it may be that access to the model is limited in some way (e.g., to registered users), but it should be possible for other researchers to have some path to reproducing or verifying the results.
        \end{enumerate}
    \end{itemize}

\item {\bf Open access to data and code}
    \item[] Question: Does the paper provide open access to the data and code, with sufficient instructions to faithfully reproduce the main experimental results, as described in supplemental material?
    \item[] Answer:  \answerYes{}
    \item[] Justification: We will publicly release the codebase and model checkpoints upon acceptance for publication, along with detailed instructions for reproducing the main experimental results. 
    \item[] Guidelines:
    \begin{itemize}
        \item The answer NA means that paper does not include experiments requiring code.
        \item Please see the NeurIPS code and data submission guidelines (\url{https://nips.cc/public/guides/CodeSubmissionPolicy}) for more details.
        \item While we encourage the release of code and data, we understand that this might not be possible, so “No” is an acceptable answer. Papers cannot be rejected simply for not including code, unless this is central to the contribution (e.g., for a new open-source benchmark).
        \item The instructions should contain the exact command and environment needed to run to reproduce the results. See the NeurIPS code and data submission guidelines (\url{https://nips.cc/public/guides/CodeSubmissionPolicy}) for more details.
        \item The authors should provide instructions on data access and preparation, including how to access the raw data, preprocessed data, intermediate data, and generated data, etc.
        \item The authors should provide scripts to reproduce all experimental results for the new proposed method and baselines. If only a subset of experiments are reproducible, they should state which ones are omitted from the script and why.
        \item At submission time, to preserve anonymity, the authors should release anonymized versions (if applicable).
        \item Providing as much information as possible in supplemental material (appended to the paper) is recommended, but including URLs to data and code is permitted.
    \end{itemize}

\item {\bf Experimental setting/details}
    \item[] Question: Does the paper specify all the training and test details (e.g., data splits, hyperparameters, how they were chosen, type of optimizer, etc.) necessary to understand the results?
    \item[] Answer: \answerYes{}
    \item[] Justification: Hyperparameter settings are reported for each experiment, and additional implementation details are provided in the Appendix B.
    \item[] Guidelines:
    \begin{itemize}
        \item The answer NA means that the paper does not include experiments.
        \item The experimental setting should be presented in the core of the paper to a level of detail that is necessary to appreciate the results and make sense of them.
        \item The full details can be provided either with the code, in appendix, or as supplemental material.
    \end{itemize}

\item {\bf Experiment statistical significance}
    \item[] Question: Does the paper report error bars suitably and correctly defined or other appropriate information about the statistical significance of the experiments?
    \item[] Answer: \answerNA{}
    \item[] Justification: We do not perform statistical significance testing in this work due to (1) the high computational cost of repeated runs and (2) the robustness of our dataset, which includes 24,000 samples.
    \item[] Guidelines:
    \begin{itemize}
        \item The answer NA means that the paper does not include experiments.
        \item The authors should answer "Yes" if the results are accompanied by error bars, confidence intervals, or statistical significance tests, at least for the experiments that support the main claims of the paper.
        \item The factors of variability that the error bars are capturing should be clearly stated (for example, train/test split, initialization, random drawing of some parameter, or overall run with given experimental conditions).
        \item The method for calculating the error bars should be explained (closed form formula, call to a library function, bootstrap, etc.)
        \item The assumptions made should be given (e.g., Normally distributed errors).
        \item It should be clear whether the error bar is the standard deviation or the standard error of the mean.
        \item It is OK to report 1-sigma error bars, but one should state it. The authors should preferably report a 2-sigma error bar than state that they have a 96\% CI, if the hypothesis of Normality of errors is not verified.
        \item For asymmetric distributions, the authors should be careful not to show in tables or figures symmetric error bars that would yield results that are out of range (e.g. negative error rates).
        \item If error bars are reported in tables or plots, The authors should explain in the text how they were calculated and reference the corresponding figures or tables in the text.
    \end{itemize}

\item {\bf Experiments compute resources}
    \item[] Question: For each experiment, does the paper provide sufficient information on the computer resources (type of compute workers, memory, time of execution) needed to reproduce the experiments?
    \item[] Answer: \answerYes{} 
    \item[] Justification: See Appendix B.
    \item[] Guidelines:
    \begin{itemize}
        \item The answer NA means that the paper does not include experiments.
        \item The paper should indicate the type of compute workers CPU or GPU, internal cluster, or cloud provider, including relevant memory and storage.
        \item The paper should provide the amount of compute required for each of the individual experimental runs as well as estimate the total compute. 
        \item The paper should disclose whether the full research project required more compute than the experiments reported in the paper (e.g., preliminary or failed experiments that didn't make it into the paper). 
    \end{itemize}
    
\item {\bf Code of ethics}
    \item[] Question: Does the research conducted in the paper conform, in every respect, with the NeurIPS Code of Ethics \url{https://neurips.cc/public/EthicsGuidelines}?
    \item[] Answer: \answerYes{}
    \item[] Justification: We have reviewed the NeurIPS Code of Ethics and confirm that our research conforms to its principles. All data used in this study are from publicly available, de-identified medical datasets, and no personally identifiable information (PII) was accessed or used.
    \item[] Guidelines:
    \begin{itemize}
        \item The answer NA means that the authors have not reviewed the NeurIPS Code of Ethics.
        \item If the authors answer No, they should explain the special circumstances that require a deviation from the Code of Ethics.
        \item The authors should make sure to preserve anonymity (e.g., if there is a special consideration due to laws or regulations in their jurisdiction).
    \end{itemize}

\item {\bf Broader impacts}
    \item[] Question: Does the paper discuss both potential positive societal impacts and negative societal impacts of the work performed?
    \item[] Answer: \answerYes{}
    \item[] Justification: See Appendix A. 
    \item[] Guidelines:
    \begin{itemize}
        \item The answer NA means that there is no societal impact of the work performed.
        \item If the authors answer NA or No, they should explain why their work has no societal impact or why the paper does not address societal impact.
        \item Examples of negative societal impacts include potential malicious or unintended uses (e.g., disinformation, generating fake profiles, surveillance), fairness considerations (e.g., deployment of technologies that could make decisions that unfairly impact specific groups), privacy considerations, and security considerations.
        \item The conference expects that many papers will be foundational research and not tied to particular applications, let alone deployments. However, if there is a direct path to any negative applications, the authors should point it out. For example, it is legitimate to point out that an improvement in the quality of generative models could be used to generate deepfakes for disinformation. On the other hand, it is not needed to point out that a generic algorithm for optimizing neural networks could enable people to train models that generate Deepfakes faster.
        \item The authors should consider possible harms that could arise when the technology is being used as intended and functioning correctly, harms that could arise when the technology is being used as intended but gives incorrect results, and harms following from (intentional or unintentional) misuse of the technology.
        \item If there are negative societal impacts, the authors could also discuss possible mitigation strategies (e.g., gated release of models, providing defenses in addition to attacks, mechanisms for monitoring misuse, mechanisms to monitor how a system learns from feedback over time, improving the efficiency and accessibility of ML).
    \end{itemize}
    
\item {\bf Safeguards}
    \item[] Question: Does the paper describe safeguards that have been put in place for responsible release of data or models that have a high risk for misuse (e.g., pretrained language models, image generators, or scraped datasets)?
    \item[] Answer: \answerYes{}
    \item[] Justification: We use only publicly available and de-identified medical datasets to ensure compliance with privacy regulations. For model release, we will provide pretrained weights under a research license that includes responsible use guidelines, including prohibitions on clinical deployment without regulatory approval.
    \item[] Guidelines:
    \begin{itemize}
        \item The answer NA means that the paper poses no such risks.
        \item Released models that have a high risk for misuse or dual-use should be released with necessary safeguards to allow for controlled use of the model, for example by requiring that users adhere to usage guidelines or restrictions to access the model or implementing safety filters. 
        \item Datasets that have been scraped from the Internet could pose safety risks. The authors should describe how they avoided releasing unsafe images.
        \item We recognize that providing effective safeguards is challenging, and many papers do not require this, but we encourage authors to take this into account and make a best faith effort.
    \end{itemize}

\item {\bf Licenses for existing assets}
    \item[] Question: Are the creators or original owners of assets (e.g., code, data, models), used in the paper, properly credited and are the license and terms of use explicitly mentioned and properly respected?
    \item[] Answer: \answerYes{}
    \item[] Justification: We use publicly available datasets and models whose licenses and sources are properly cited in the paper.
    \item[] Guidelines:
    \begin{itemize}
        \item The answer NA means that the paper does not use existing assets.
        \item The authors should cite the original paper that produced the code package or dataset.
        \item The authors should state which version of the asset is used and, if possible, include a URL.
        \item The name of the license (e.g., CC-BY 4.0) should be included for each asset.
        \item For scraped data from a particular source (e.g., website), the copyright and terms of service of that source should be provided.
        \item If assets are released, the license, copyright information, and terms of use in the package should be provided. For popular datasets, \url{paperswithcode.com/datasets} has curated licenses for some datasets. Their licensing guide can help determine the license of a dataset.
        \item For existing datasets that are re-packaged, both the original license and the license of the derived asset (if it has changed) should be provided.
        \item If this information is not available online, the authors are encouraged to reach out to the asset's creators.
    \end{itemize}

\item {\bf New assets}
    \item[] Question: Are new assets introduced in the paper well documented and is the documentation provided alongside the assets?
    \item[] Answer: \answerYes{}
    \item[] Justification: We introduce several new assets as part of this work, including pretrained MedVista3D model checkpoints, LLM-rewritten radiology reports for CT-RATE. These assets will be released with accompanying documentation upon paper acceptance.
    \item[] Guidelines:
    \begin{itemize}
        \item The answer NA means that the paper does not release new assets.
        \item Researchers should communicate the details of the dataset/code/model as part of their submissions via structured templates. This includes details about training, license, limitations, etc. 
        \item The paper should discuss whether and how consent was obtained from people whose asset is used.
        \item At submission time, remember to anonymize your assets (if applicable). You can either create an anonymized URL or include an anonymized zip file.
    \end{itemize}

\item {\bf Crowdsourcing and research with human subjects}
    \item[] Question: For crowdsourcing experiments and research with human subjects, does the paper include the full text of instructions given to participants and screenshots, if applicable, as well as details about compensation (if any)? 
    \item[] Answer: \answerNA{}
    \item[] Justification: This paper does not involve crowdsourcing or research with human subjects. All data used are from publicly available, de-identified medical datasets with appropriate licenses and do not involve any direct interaction with individuals.
    \item[] Guidelines:
    \begin{itemize}
        \item The answer NA means that the paper does not involve crowdsourcing nor research with human subjects.
        \item Including this information in the supplemental material is fine, but if the main contribution of the paper involves human subjects, then as much detail as possible should be included in the main paper. 
        \item According to the NeurIPS Code of Ethics, workers involved in data collection, curation, or other labor should be paid at least the minimum wage in the country of the data collector. 
    \end{itemize}

\item {\bf Institutional review board (IRB) approvals or equivalent for research with human subjects}
    \item[] Question: Does the paper describe potential risks incurred by study participants, whether such risks were disclosed to the subjects, and whether Institutional Review Board (IRB) approvals (or an equivalent approval/review based on the requirements of your country or institution) were obtained?
    \item[] Answer: \answerNA{}
    \item[] Justification: This study does not involve human subjects or direct human participation. Therefore, IRB approval was not required.
    \item[] Guidelines:
    \begin{itemize}
        \item The answer NA means that the paper does not involve crowdsourcing nor research with human subjects.
        \item Depending on the country in which research is conducted, IRB approval (or equivalent) may be required for any human subjects research. If you obtained IRB approval, you should clearly state this in the paper. 
        \item We recognize that the procedures for this may vary significantly between institutions and locations, and we expect authors to adhere to the NeurIPS Code of Ethics and the guidelines for their institution. 
        \item For initial submissions, do not include any information that would break anonymity (if applicable), such as the institution conducting the review.
    \end{itemize}

\item {\bf Declaration of LLM usage}
    \item[] Question: Does the paper describe the usage of LLMs if it is an important, original, or non-standard component of the core methods in this research? Note that if the LLM is used only for writing, editing, or formatting purposes and does not impact the core methodology, scientific rigorousness, or originality of the research, declaration is not required.
    \item[] Answer: \answerYes{}
    \item[] Justification: Large language models such as GPT-4o and Qwen2.5, were used to rewrite radiology reports for improving semantic clarity during pretraining. See method section and Figure~\ref{fig:fig1}.
    \item[] Guidelines:
    \begin{itemize}
        \item The answer NA means that the core method development in this research does not involve LLMs as any important, original, or non-standard components.
        \item Please refer to our LLM policy (\url{https://neurips.cc/Conferences/2025/LLM}) for what should or should not be described.
    \end{itemize}

\end{enumerate}


\appendix
\section{Broader Impact}
\paragraph{Reproducibility statement.} We are committed to efficient and reproducible research. Our code and datasets will be publicly released.
\paragraph{Potential benefits.} MedVista3D could support radiologists by improving diagnostic accuracy, automating report generation, and medical image segmentation.
\paragraph{Potential risks.} Using large language models for rewriting medical reports may inadvertently introduce hallucinated content if not properly validated. While our approach is intended for training supervision, not clinical deployment, future work should explore robust validation pipelines to detect hallucinations, ensure factual consistency, and maintain clinical reliability. We emphasize that such systems should only be used in practice with rigorous testing, explainability safeguards, and alignment with domain expertise.

\section{Implementation details}
\textbf{Pretraining on CT-RATE:} For dataset preprocessing, we use volumes resampled to 3.0 mm $\times$ 1.0 mm $\times$ 1.0 mm from Radgenome dataset \cite{radgenome}. We also use its segmentation masks and region sentences for regional image-text alignment. For intensity normalization, we follow CT-CLIP \cite{hamamci2024developing} preprocessing. We uniformly resize all volumes to 96 $\times$ 320 $\times$ 320 using padding or center cropping. For text encoder, we use BiomedVLP-CXR-BERT-specialized \cite{boecking2022making}. For vision encoder, we use 1). ViT-B, with embedding dimension as 768 and depth as 12; and 2). UniMISS-Small \cite{unimiss}. We train MedVista3D-ViT using a batch size of 32 and MedVista3D-UniMISS using a batch size of 20 for a total of 16 epochs on our proposed loss. The optimizer is AdamW with the weight decay of 1e-5. We use a linear warmup with cosine decay scheduler for 200 steps and a learning rate of 5e-5. All experiments were conducted using NVIDIA A100 GPUs (80GB) on an internal cluster.

\textbf{Visual question answering using MedVista3D-LLaVA:} We use Llama-3.1-7B \cite{llama3} as the language decoder and the pretrained MedVista3D-ViT as the vision encoder. For multi-modal projector, we use a two-layer MLP-GELU following LLaVA-1.5 \cite{llava}. We follow the two-stage training strategy same as LLaVA: 1). First, we perform contrastive alignment using CT-RATE's volume-report pairs to tune the multi-modal projector; 2). Second, we perform supervised finetuning using LoRA \cite{hu2021lora} with rank $r$ set to 128, scaling factor $\alpha$ set to 256, and a learning rate of 2e-5. We train a total of 10 epochs following CT-CHAT. 

\textbf{Segmentation on TotalSegmentator:} We adopt the pretrained MedVista3D-UniMISS as the segmentation encoder. We attach STU-Net-B's decoder \cite{huang2023stu} to our encoder. We use nn-UNet \cite{isensee2021nnu} to preprocess the TotalSegmentator dataset and train within their framework. We use a learning rate of 5e-5 and a batch size of 2. Input volumes are uniformly cropped to 128 $\times$ 128 $\times$ 128. We train for a total of 1000 epochs following the default setting.

\textbf{Classification on STOIC 2021:} MedVista3D-UniMISS is initialized with pretrained CT-RATE weights. We resample CT volumes to 3.0 mm $\times$ 1.0 mm $\times$ 1.0 mm and crop/pad to 96 $\times$ 320 $\times$ 320. From the full 2000 volumes, we randomly select 80\% for training, 10\% for validation and 10\% for testing. We use a batch size of 96, a learning rate of 1e-4 and finetune for 10 epochs.

\section{Global vs Local Alignment}
\textbf{Global alignment.} Contrastive VLM aims to learn positive-negative image-text embeddings by jointly training an image encoder \( f_{\text{I}}(\cdot) \) and a text encoder \( f_{\text{T}}(\cdot) \). One common approach is global image-text alignment, such as CLIP \cite{radford2021learning}. Given a dataset of \( P \) pairs of CT image volumes and their corresponding radiology reports, \( X = \{x_1, \dots, x_P \} \) and \( Y = \{y_1, \dots, y_P\} \), the global embeddings for the \( i \)th volume-report pair can be obtained as
\( v^G_i = f_{\text{I}}(x_i) \) and \( t^G_i = f_{\text{T}}(y_i) \), where \( x_i \in \mathbb{R}^{1 \times D \times H \times W} \) and \( y_i \in \mathbb{R}^{l} \) represent the dimensions of the input CT volume and radiology report, respectively.
To align image and text representations, a contrastive objective pushes the embeddings of matched volume-report pairs together while pushing those of unmatched pairs apart. Using InfoNCE loss \cite{oord2018representation}, the global alignment objective becomes, 
\begin{align}
\mathcal{L}_{\text{Global}} &= \frac{1}{2} \left[\mathcal{L}^{G}_{I \to T} + \mathcal{L}^{G}_{T \to I}\right], \label{eq:dual_global}
\end{align}
The first term consists of the global image-to-text loss, $\mathcal{L}_{I\to T}^G$, and is defined as, 
\begin{align}
\mathcal{L}^{G}_{I \to T} &= -\frac{1}{N} \sum_{i=1}^N 
\log \frac{\exp \big( \text{sim}(v_i^{G}, t_i^{G}) / \tau \big)}{\sum_{j=1}^N \exp \big( \text{sim}(v_i^{G}, t_j^{G}) / \tau \big)}, \label{eq:2}
\end{align}
where $N$ is the batch size and $\text{sim}(\cdot, \cdot)$ is the similarity function and $\tau$ is a learnable logit. We omit ${L}^{G}_{T \to I}$ since it is symmetric. 

\textbf{Local alignment.} However, global approach can overlook fine-grained similarities or differences among various organs. Alternatively, local image-text alignment identifies all possible regions in the CT image and extracts region-specific features \cite{fVLM, ct-glip}. Assuming the CT image can be divided into image regions $x^{1}_{i}, \dots, x^{r}_{i}$, and radiology reports can also be decomposed into fine-grained captions $y^{1}_{i}, \dots, y^{r}_{i}$ describing each organ, region-text pairs can be formed as $\{(x^{1}_{i}, y^{1}_{i}), \dots, (x^{r}_{i}, y^{r}_{i})\}$. For region $r$, $f_{I}$ extracts local image embedding $v_i^{r}$ and $f_{T}$ extracts local text embeddings $t_i^{L}$. The local alignment loss can be defined as: 
\begin{align}
\mathcal{L}_{\text{Local}} &= \frac{1}{2} \left[\mathcal{L}^{L}_{I \to T} + \mathcal{L}^{L}_{T \to I} \right], \label{eq:3}
\end{align}

The local image-to-text loss can be written as:
\begin{align}
\mathcal{L}^{L}_{I \to T} = -\frac{1}{RN} \sum_{r=1}^R \sum_{i=1}^N 
\log \frac{\exp \Big( \text{sim}\big(v_i^{r}, t_i^{r})\big) / \tau \Big)}
{\sum_{j=1}^N \exp \Big( \text{sim}\big(v_i^{r}, t_j^{r}\big) / \tau \Big)}, \label{eq:4}
\end{align}
where $R$ is the total number of regions. However, local alignment methods often lack broader contextual information and requires anatomical priors (i.e. segmentation masks), which may not always be feasible in clinical settings. 

\section{Prompting LLMs for improving disease semantics}
\begin{figure*}[ht] 
    \centering
    \includegraphics[width=0.7\textwidth]
    {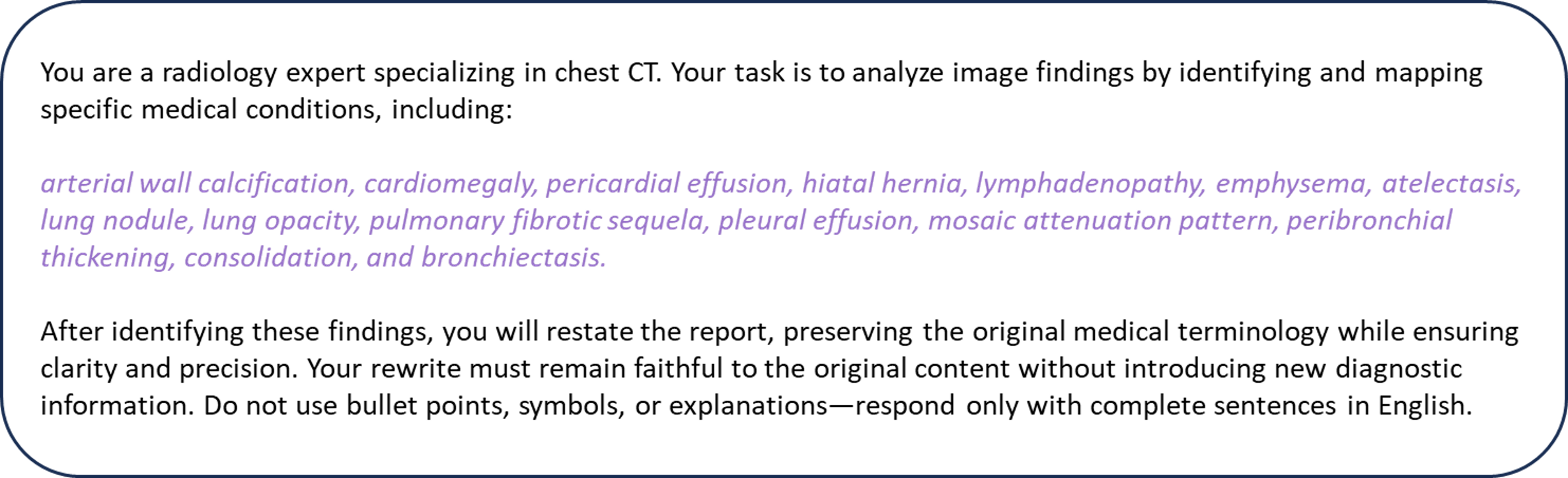} 
    \caption{Prompts for report-level rewrites to emphasize disease presences.}
    \label{fig:fig5}
\end{figure*}

\begin{figure*}[ht] 
    \centering
    \includegraphics[width=0.7\textwidth]
    {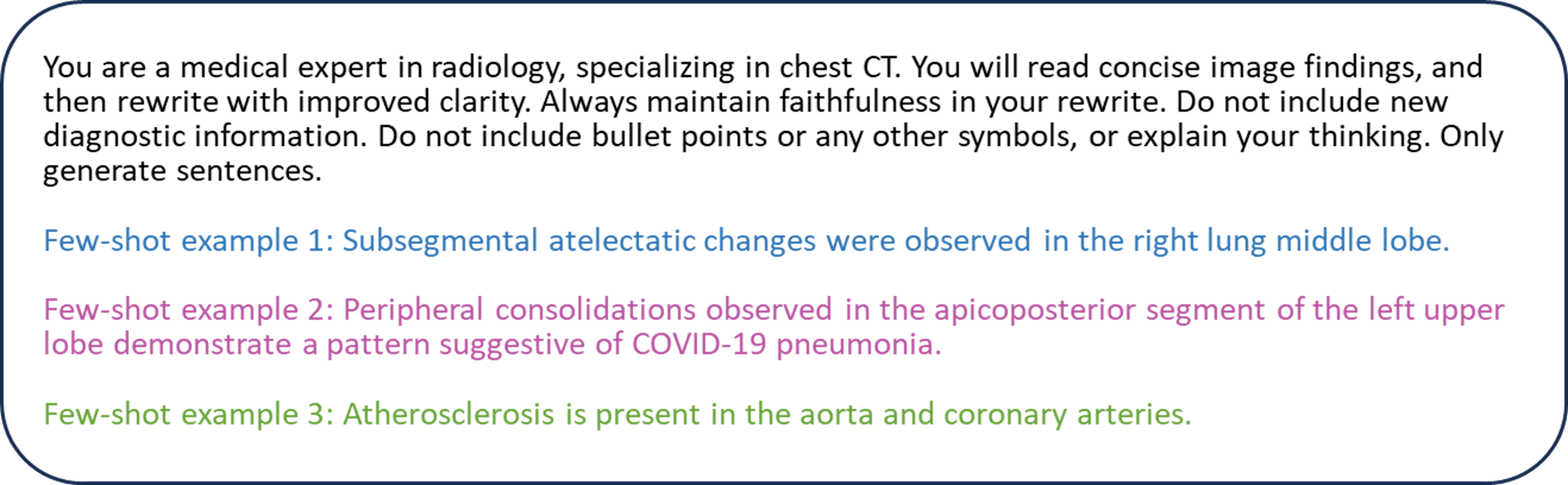} 
    \caption{Prompts for region-level rewrite with few-shot prompting.}
\end{figure*}
\newpage
\section{Training algorithm}
The training pseudo code of MedVista3D is shown below:
\begin{algorithm}
\caption{MedVista3D}
\begin{algorithmic}[1]
\Require $ (x_{i}^{G}, y_{i}^{G})_{i=1}^B, (x_{i}^{r}, y_{i}^{r})_{i=1}^B, (x_{i}^{G}, \hat{y}_{i}^{G})_{i=1}^B, (x_{i}^{r}, \hat{y}_{i}^{r})_{i=1}^B, f_{I}, f_{T}, \text{RSMB}$
\Function{Compute\_MedVista3D\_Loss}{$f_{I}, f_{T}$}

    \State $v_{i}^{G}, t_{i}^{G} \gets f_{I}(x_{i}^{G}), f_{T}(y_{i}^{G})$
    \Comment{Global features.}
    \State $v_{i}^{r}, t_{i}^{r} \gets f_{I}(x_{i}^{r}), f_{T}(y_{i}^{r})$
    \Comment{Local features.}
    \State $v_{i}^{G}, \hat{t}_{i}^{G} \gets f_{I}(x_{i}^{G}), f_{T}(\hat{y}_{i}^{G})$
    \Comment{Global semantic features.}
    \State $v_{i}^{r}, \hat{t}_{i}^{r} \gets f_{I}(x_{i}^{r}), f_{T}(\hat{y}_{i}^{r})$
    \Comment{Local semantic features.}

    \Statex
    \State $\hat{t}_{i}^{G_{NN}} \gets \text{Top-1 nearest neighbor of } \hat{t}_{i}^{G} \text{ from RSMB}$
    \Comment{Global semantic query with RSMB.}
    \State $\hat{t}_{i}^{{r}_{NN}} \gets \text{Top-1 nearest neighbor of } \hat{t}_{i}^{r} \text{ from RSMB}$
    \Comment{Local semantic query with RSMB.}
    \Statex
    \State \text{Compute} $L_{\text{Global}} \text{ from } (v_{i}^{G}, t_{i}^{G})$ and $L_{\text{Local}} \text{ from } (v_{i}^{r}, t_{i}^{r})$
    \State \text{Compute} $L_{\text{Global Semantic}} \text{ from } (v_{i}^{G}, \hat{t}_{i}^{{G}_{NN}})$ and $L_{\text{Local Semantic}} \text{ from } (v_{i}^{r}, \hat{t}_{i}^{{r}_{NN}})$
    \State \text{Compute} $L_{\text{MedVista3D}} \text{ from } L_{\text{Global}},   L_{\text{Local}}, L_{\text{Global Semantic}} \text{ and } L_{\text{Local Semantic.}}$
    
    \Comment{Calculate the losses.}
    \State $\text{Backward } L_{\text{MedVista3D}} \text{ and update } f_{I}, f_{T} $
    \Comment{Update the network.}
    \Statex
    \State $\text{RSMB} \gets \text{Queue\_Update}(\text{RSMB}, \hat{t}_{i}^{{G}})$
    \State $\text{RSMB} \gets \text{Queue\_Update}(\text{RSMB}, \hat{t}_{i}^{{r}})$
    \Comment{Update RSMB.}

\EndFunction
\\
\Function{Queue\_Update}{$\text{RSMB}, \hat{t}_{i}$}
    \State $B \gets \text{batch size of } \hat{t}_{i}$ 
    \State $ptr \gets \text{next free position in RSMB}$ 
    \State $S \gets \text{length of RSMB}$
    \If{$ptr + B \geq S$} 
        \Comment{Queue size is exceeded.}
        \State $RSMB[:, ptr:S] \gets \hat{t}_{i}[:, 0:(S - ptr)]$
        \Comment{Fill remaining slots.}
        \State $ptr \gets 0$
        \Comment{Reset pointer to the start.}
    \Else
        \State $RSMB[:, ptr:ptr + B] \gets \hat{t}_{i}$
        \Comment{Push embeddings into the queue.}
        \State $ptr \gets ptr + B$
        \Comment{Advance pointer by the batch size.}
    \EndIf
    
    \State \Return $\text{RSMB}$ 
    \Comment{Return updated RSMB.}
\EndFunction
\end{algorithmic}
\end{algorithm}
\end{document}